\documentclass[runningheads]{llncs}

\usepackage{latexsym}
\usepackage{graphicx}
\usepackage{url}
\usepackage{enumerate}
\usepackage{color}
\usepackage{subfigure}
\usepackage{multirow}
\usepackage{xcolor}
\usepackage[sort]{cite} 
\usepackage{mathtools}

\begin{document}

\title{Evaluating Perceived Usefulness and Ease of Use of CMMN and DCR}

\author{Amin Jalali}

\titlerunning{Evaluating Perceived Usefulness and Ease of Use of CMMN and DCR}

\institute{Department of Computer and Systems Sciences, \\Stockholm University, \\Sweden\\
\{aj\}@dsv.su.se}

\maketitle

\begin{abstract}

Case Management has been gradually evolving to support Knowledge-intensive business process management, which resulted in developing different modeling languages, e.g., Declare, Dynamic Condition Response (DCR), and Case Management Model and Notation (CMMN).
A language will die if users do not accept and use it in practice - similar to extinct human languages. 
Thus, it is important to evaluate how users perceive languages to determine if there is a need for improvement. 
Although some studies have investigated how the process designers perceived Declare and DCR, there is a lack of research on how they perceive CMMN. 
Therefore, this study investigates how the process designers perceive the usefulness and ease of use of CMMN and DCR based on the Technology Acceptance Model. 
DCR is included to enable comparing the study result with previous ones.
The study is performed by educating master level students with these languages over eight weeks by giving feedback on their assignments to reduce perceptions biases. 
The students' perceptions are collected through questionnaires before and after sending feedback on their final practice in the exam. 
Thus, the result shows the perception of participants can change slightly by receiving feedback, while the change is not significant due to being well trained.
The reliability of responses is tested using Cronbach's alpha, and the result indicates that both languages have an acceptable level for both perceived usefulness and ease of use.

\vspace{+6pt}
\textbf{Keywords:} business process modeling, knowledge-intensive, case management, CMMN, DCR

\end{abstract}

\section{Introduction}\label{Sec:Introduction}

Case Management is a research paradigm that supports knowledge-intensive process (KiP) management~\cite{di2015knowledge}.
The support is defined around the concept of the case, e.g., patient in the healthcare or customer in the insurance domain.
In these processes, knowledge workers decide how a case shall proceed instead of pre-defined rigid rules.
As a result, the support for KiPs differs from the workflow-based processes so that even the management lifecycle needs its variation, known as Collaborative knowledge work lifecycle~\cite{di2015knowledge}.

The collaborative nature of work in KiPs requires more freedom by knowledge workers to decide how a case shall proceed.
Therefore, traditional workflow-based models could not support case management, for they become too complex by capturing the high degree of flexibilities required by different cases.
Thus, several case management modeling languagues are defined, e.g., Declare~\cite{pesic2008constraint,pesic2007declare}, Dynamic Condition Response (DCR)~\cite{hildebrandt2011declarative}, and Case Management Model and Notation (CMMN)~\cite{cmmn} which is based on Guard-Stage-Milestone (GSM)~\cite{hull2011business}.
As these languages are new, it is hard to predict if they will be accepted by users, which opens up rooms for further investigation.

The user acceptance of process modeling languages, like other information systems, can be evaluated based on two variables, i.e., perceived usefulness (PU) and perceived ease of use (PEU)~\cite{davis1989perceived}.
Currently, there are few studies on how users perceive Declare and DCR, e.g.~\cite{zugal2015investigating,haisjackl2016understanding,reijers2013declarative}.
However, there is a gap in how users perceive CMMN.
Therefore, this study aims to evaluate the user acceptance of CMMN language. 
It also includes DCR to have a comparison reference to relate the result to other investigations.
DCR is selected because it is expected to be more comprehensive than Declare, as it has fewer modeling elements~\cite{reijers2013declarative}.
The evaluation is performed based on the Technology Acceptance Model~\cite{davis1989perceived}.

As these languages are new, it is hard, if not impossible, to find users who already know them. 
Thus, the users shall be trained, and the acceptance shall be measured based on their perception (self-assessment).
As the self-assessment is a subjective score, it can differ during the learning process due to biased factors, i.e., over- or under-confidence~\cite{eberlein2011effects}.
These biased factors can be minimized by repeated experiences and feedback~\cite{eberlein2011effects}.
In this study, we aim to answer this research question: 
\begin{itemize}
	\item \textit{How do trained process designers perceive the usefulness and the ease of use of DCR and CMMN languages for modeling knowledge-intensive processes?}
\end{itemize}

To answer this question, we trained students in the business process and case management course at Stockholm University.
The students practiced DCR and CMMN for around eight weeks and received feedback on their assignments - to minimize the overconfidence and underconfidence biases. 
Finally, we collected perceptions from those who were interested in participating in this study before and after the feedback on their performance on the exam.
Participation in this study was optional.
The reliability of responses is tested using Cronbach's alpha, and the result shows that both languages have an acceptable level for both perceived usefulness and ease of use.

The remainder of this paper is structured as follows. 
Section~\ref{Sec:Background} gives a brief background on related work.
Section~\ref{Sec:Method} describes the method that is used in this study.
Section~\ref{Sec:Result} reports the result, and 
Section~\ref{Sec:Conclusion} concludes the paper.

\section{Background}\label{Sec:Background}

This section summarizes related works and excerpts of DCR and CMMN notations, which are used later to present sample processes.
Please note that we do not aim to give the full syntax of these languages, which can be found in related literature.

\subsection{Users perceptions in business process management}

A language will eventually die if people do not accept and use it in practice, which is also true for business process modeling languages.
Thus, it is important to evaluate how users perceive languages to determine if there is a need for improvement. 
The evaluation can help us to improve process modeling languages.
Here, we mention some related work that evaluated the user acceptance in the Business Process Management (BPM) area, in general, and in the case management area, in particular.

\subsubsection{Users perceptions in business process modeling\\}
Process models can easily become complex as they represent complex business processes in practice, so different approaches have been developed to enable process designers to deal with the complexity.
La Rosa et al., categorize these approaches into two main category, i.e., concrete syntax modifications~\cite{la2011managingconcrete} and abstract syntax modifications~\cite{la2011managingabstract}.
They also identified different patterns that can be applied in each category.

The concrete syntax modifications refer to i) using highlights, ii) following layout guidelines, iii) following naming guidelines, or iv) applying different representations techniques, e.g., using icons for tasks or etc~\cite{la2011managingconcrete}.
The abstract syntax modifications refer to applying different abstraction techniques in business process modeling, e.g., using vertical, horizontal, or orthogonal modularization technqiues~\cite{la2011managingabstract}.
La Rosa et al. evaluated how users perceived usefulness and ease of use of the identified patterns by applying the technology acceptance model~\cite{davis1989perceived}.
Their evaluation study showed that all identified patterns are perceived as relevant.

The users' perception evaluation is important because the artifact's actual usage is influenced by the potential users' perceptions - which can be measured in terms of usefulness and ease of use~\cite{davis1989perceived}.
Therefore, researchers have used techniques like the technology acceptance model to evaluate different business process modeling techniques.
For example, the technology acceptance model is used to evaluate 
i) how users perceived orthogonal modularization based on aspect-oriented business process modeling in~\cite{jalali2018hybrid,jalali2018weaving}, 
ii) how users perceived the vertical decomposition using BPMN in~\cite{turetken2019influence}, and
iii) how graphical highlights can increase the cognitive effectiveness of business process models in~\cite{jovst2017improving}.

\subsubsection{Users perceptions in case management\\}
Case management is fairy new paradigm in comparison with Business Process Management, and few languages have been developed to support managing cases.
Declarative Service Flow Language (DecSerFlow)~\cite{van2006decserflow} (a.k.a., Declare) can be considered as one of the first attempts for such lanaguagues.

The understandability and maintainability of process models is a crucial requirement for any process modeling language.
Thus, Fahland et al. identified and hypothesized a set of propositions for differences between imperative and declarative process modeling languages concerning understandability and maintainability in \cite{fahland2009declarative_understandability} and \cite{fahland2009declarative_maintanability}, respectively.
Weber et al.~\cite{weber2009declarative} conducted a controlled experiment to investigate if process designers can deal with increased levels of constraints when using Declare.
The participants were 41 students from two universities.
The results show that the participants can deal with introduced constraints, which justified further development of declarative process modeling.

Pichler et al. investigated if the imperative or the declarative process modeling languages are better understood by running an experiment with 27 students~\cite{pichler2011imperative}.
Their study shows that the imperative process modeling languages are better understood by students.
The technology acceptance model is used for the first time to assess how professionals perceived Declare, and DCR languages by Reijers et al.~\cite{reijers2013declarative}.
Ten professionals from the industry participated in their workshop, and the result shows that they found the languages easy to learn.
This study also revealed the potential for defining hybrid approaches.
Several hybrid process modeling is proposed, among which the understandability of DCR-HR is investigated through an explorative study~\cite{andaloussi2019exploring}.

Zugal et al.~\cite{zugal2015investigating} investigated the effect of using hierarchies in expressiveness and understandability of declarative models.
The study is based on nine participants in two universities.
It shows that hierarchies enhance expressiveness.
It also shows that the hierarchies can increase the models' understandability, but they should be applied with care.
Haisjackl et al.~\cite{haisjackl2016understanding} investigated how declarative models are understood through an explorative study.
Their study shows that the subjects could understand a single constraint well, but it was challenging for them to handle a combination of constraints.
This study also shows that some graphical notation in Declare, which are similar to imperative modeling languages, causes considerable trouble in understandability.

\subsection{Dynamic Condition Response (DCR)}

Hildebrandt and Mukkamala introduced DCR in 2010 as a declarative process modeling languague~\cite{hildebrandt2010distributed}.
The syntax of the language includes the definition of nodes (the group of an activity and its roles) and the relation that can be defined among them, i.e., response ($\mathrel{\bullet}\joinrel\rightarrow$), condition ($\rightarrow\joinrel\mathrel{\bullet}$), inclusion ($\rightarrow\joinrel\mathrel{+}$), and exclusion ($\rightarrow\joinrel\mathrel{\%}$)~\cite{hildebrandt2010distributed}.

They also defined semantics for DCR, where several events can occur for a node.
A node can be included or excluded from the process structure.
A node can also be in the pending state, meaning that the process cannot successfully be finished until an event of the node occurs.
In short, the response relation among nodes a and b ($a\mathrel{\bullet}\joinrel\rightarrow b$) means that the state of node b will be pending if an event of node a happens. 
More precisely, event b  must eventually happen if event a happens.
The condition relation among nodes a and b ($a\rightarrow\joinrel\mathrel{\bullet}b$) means that an event of b cannot be occured unless a occurs.

The DCR's syntax and semantics have evolved over the years.
For example, the syntax is enriched to represent the excluded nodes by dashed border line~\cite{hildebrandt2011declarative}.
The extended syntax also represents a node's pending state by decorating it with an exclamation mark ($!$).
In addition, milestone relation ($\rightarrow\joinrel\mathrel{\diamond}$) and nested nodes are also introduced to the languague~\cite{hildebrandt2011nested}.
The milestone relation among two nodes a and b ($a\rightarrow\joinrel\mathrel{\diamond}b$) means that b can occur as long as a is not in the pending state.

\begin{figure*}[t!]
	\vspace{-\baselineskip}
	\begin{center}
		\includegraphics[width=1\textwidth]{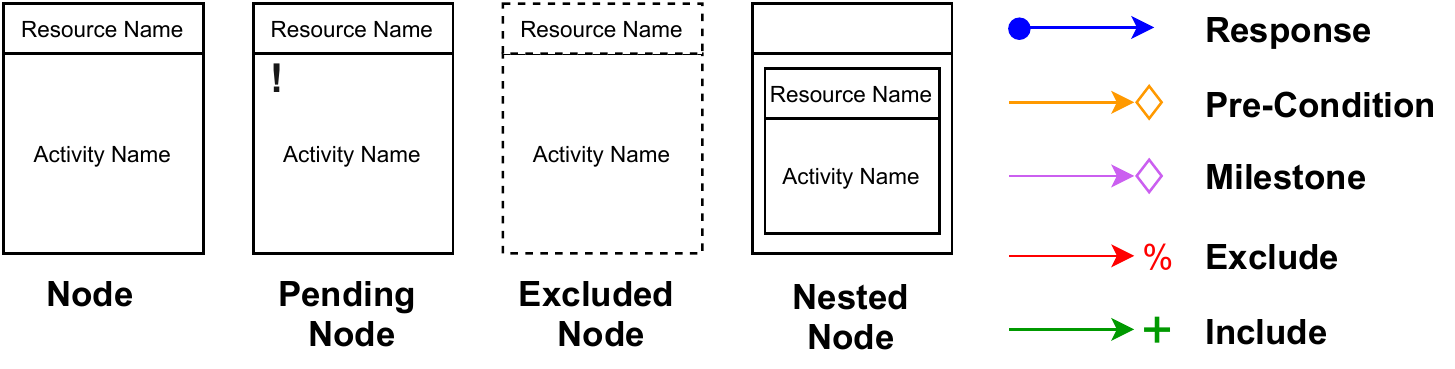}
		\caption{An excerpt of DCR Syntax}\label{Fig:DCR_Legends}
		\vspace{-2\baselineskip}
	\end{center}
\end{figure*}

DCR graph is the toolset that enables modeling and simulation of DCR models.
Currently, it supports other relations, including Pre-Condition.
The pre-condition relation among two nodes a and b means that $a\rightarrow\joinrel\mathrel{\diamond}b \wedge a\rightarrow\joinrel\mathrel{\bullet}b$.
This relation is represented with the same graphical notation as the milestone but with a different color.
\figurename~\ref{Fig:DCR_Legends} shows an excerpt of DCR syntax.

\subsection{Case Management Model and Notation (CMMN)}

\begin{figure*}[b!]
	\vspace{-\baselineskip}
	\begin{center}
		\includegraphics[width=0.9\textwidth]{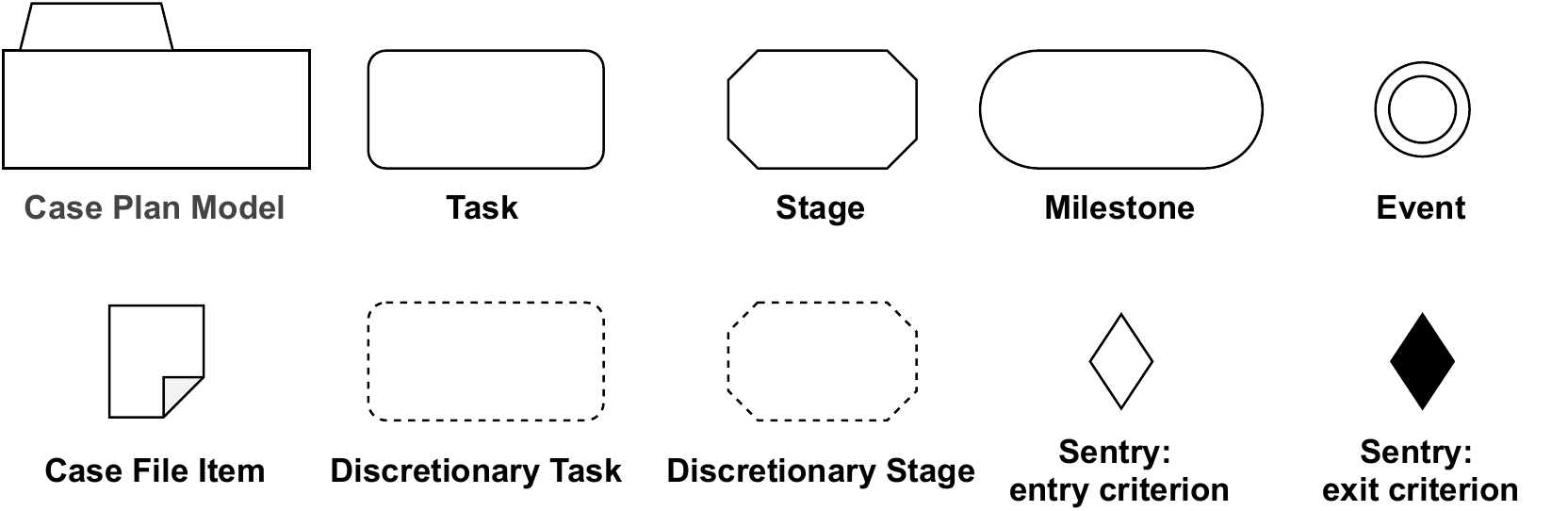}
		\caption{CMMN Legends - basic elements}\label{Fig:CMMN_Legends}
		\vspace{-2\baselineskip}
	\end{center}
\end{figure*}

Case Management Model and Notation (CMMN) is a case modeling language which is defined by Object Management Group (OMG)~\cite{cmmn}.
This language is developed by extending the Guard-Stage-Milestone (GSM) languague~\cite{hull2010introducing,hull2011business}.
\figurename~\ref{Fig:CMMN_Legends} shows an excerpt of CMMN syntax.

The \textit{case plan model}, represented by a folder, is the core part of a CMMN model.
The \textit{case plan model} captures the complete behavior of a case, and all other elements will be children of the \textit{case plan model}.
The \textit{case file item} represents the data, \textit{Task} represents activities that can happen, \textit{Stage} represents a container that includes other elements (like subprocess in BPMN), \textit{Milestone} represents an achievable target, and \textit{event} respresents something that happens during the course of a case.
Some elements like tasks can have sentries on their border.
A \textit{sentry} can represent \textit{entry criterion} or \textit{exit criterion}, which defines the condition  or event - based on which the element can be enabled or terminated, respectively.

Tasks and Stages can also be represented by the dashed borderline, which is known as \textit{discretionary task} or \textit{discretionary stage}.
Discretionary elements are not available to knowledge workers at runtime.
However, they can add these elements to their case plan at runtime.
These elements can be related to each other through lines, and the connection rules are defined in CMMN specification~\cite{cmmn}.
Elements can also be decorated using different decoration icons. 
For example, $!$ or $\#$ indicates that the task, stage, or milestone are mandatory or repetitive, respectively.
\section{Research Method}\label{Sec:Method}

Davis F.D. describes how the actual system's usage depends on the attitude of users, which can be predicted using two variables: perceived usefulness (PU) and perceived ease of use (PEU) of a system~\cite{davis1985technology}.
He also defined measurement scales that can be used to evaluate these variables~\cite{davis1989perceived}.
This evaluation technique is widely used to evaluate different information systems including different business process modeling languagues, e.g.~\cite{la2011managingconcrete,la2011managingabstract,jalali2018weaving,jalali2018hybrid,reijers2013declarative}.
This study adopted the technology acceptance model to evaluate how users perceive DCR and CMMN in concern with these variables.

The user acceptance evaluation is a sort of subjective score because users need to respond based on self-assessment.
Thus, the result can change during repeated experiences due to self-assessment biases.
These biases are rooted in over- or under-confidence, which can be minimized by repeated experience and feedback~\cite{eberlein2011effects}.
Indeed, the answers will be more reliable when these biases are minimized.

\begin{figure*}[b!]
	\vspace{-\baselineskip}
	\begin{center}
		\includegraphics[width=1\textwidth]{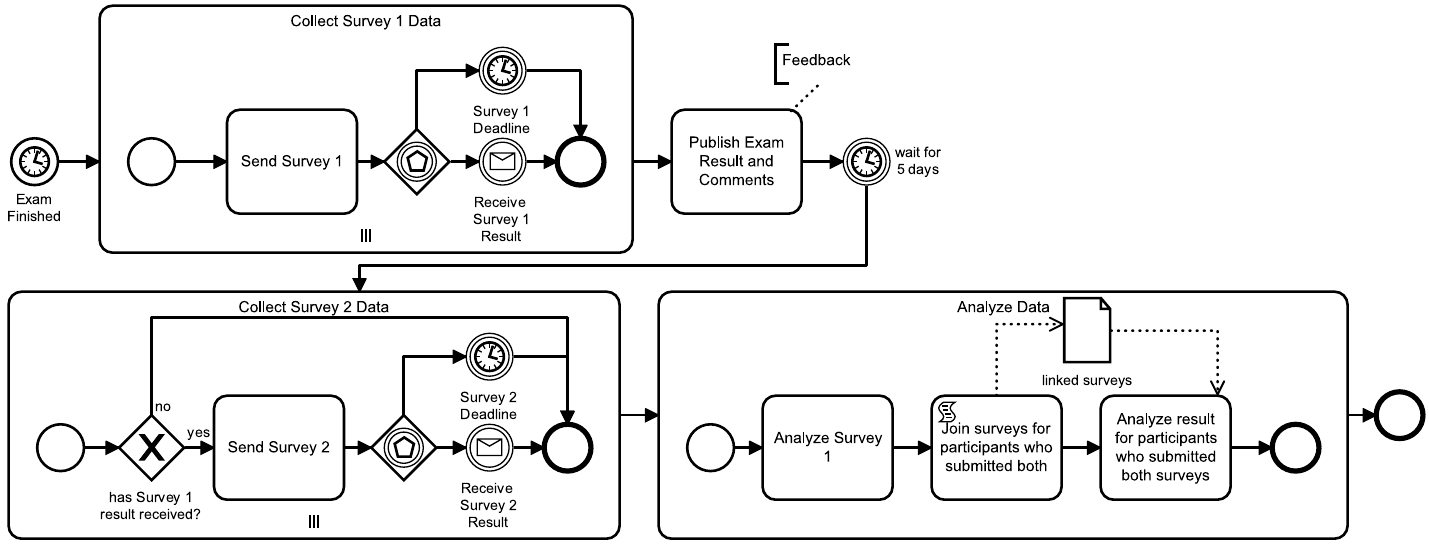}
		\caption{Research Steps}\label{Fig:method} 
		\vspace{-2\baselineskip}
	\end{center}
\end{figure*}

To minimize the over- and under-confidence, we trained the students in the business process and case management course at Stockholm University for around eight weeks.
In this period, students practiced DCR and CMMN languages in groups by designing process models, and they received feedback.
They also had two sessions with an external expert to ask their questions for DCR language.
Then, they participated in the exam where they needed to design a DCR and a CMMN model for a given case description individually.
The data collection and data processing for this study starts after the exam, which is shown in \figurename~\ref{Fig:method}.
Participation in this study was voluntary.

In this study, we collect data from students through two surveys conducted before and after giving feedback on their exams.
We call them Survey 1 and Survey 2, respectively.
The surveys were identical, but students did not know about it beforehand.
The questions are defined based on~\cite{davis1985technology}, where participants could response by choosing options in the range of 
\textit{extremely unlikely} (1),
\textit{quite unlikely} (2), 
\textit{slightly unlikely} (3), 
\textit{neither likely nor unlikely} (4), 
\textit{slightly likely} (5), 
\textit{quite likely} (6), and 
\textit{extremely likely} (7).

After the exam, survey 1 is sent out, and the responses are collected upon the announced deadline.
Then, we published the grades and comments for the exam result.
We gave students five days to go through the comments and discuss their questions with the teacher.
Then, we sent out the second survey to those who participated in the first survey.
We collected data that were submitted before the announced deadline.
Finally, we started data analysis.

We analyzed the collected data from Survey 1, and we name it \textit{Study 1}.
The second survey responses are linked with the first one to analyze the result before and after the feedback.
We analyzed the collected data from the linked data source containing both Survey 1 and Survey 2, and we name it \textit{Study 2}.
This study includes the result of students who participated in both Survey 1 and Survey 2, enabling us to track how opinions are changed after receiving the feedback.
We tested if the changes in responses are significantly different through three nonparametric test of the null hypothesis techniques.
To test the reliability of responses, we used Cronbach's alpha, which is widely used in related work, e.g.~\cite{palm2006determinants,masrom2007technology,arachchilage2013game,chae2002information}.
The Cronbach's alpha value above 0.7 is usually considered as reliable.

The details of the study will be given in the result section. Here, we also give the case description for which students designed a DCR and a CMMN model.

\subsection{Case Description}

Managing courses is a sort of knowledge-intensive process which relies on the skills and knowledge of academic staff as knowledge workers. This text describes the course management process at the department of computer and systems sciences (DSV) at Stockholm University.

The process starts when the course director registers each course coordinator when the course planning period begins. After the course coordinator is registered, (s)he can set up the courses. (S)he must publish the course contents and define the course schedule. It is important to check potential conflicts among mandatory sessions among different courses. Thus, the administrative personnel needs to check the conflicts after the course coordinator schedule the course.  If they approve, the course coordinator needs to release the course, so students can see the schedule. Indeed, the course coordinator shall release the course after (s)he apply any changes so that students can see the changes. 

To avoid having unreleased courses, the administrative personnel will notify course coordinators sometimes before the academic term starts. Note that the course coordinator can change the course content and schedule during the course several times, but the same process shall be followed. After the course is released, the course coordinator can run the course, which includes Registering groups, Publishing Recorded Lectures, opening submission box, and Registering Assignments Grades. These activities can be done several times during the course. 

The course coordinator can also start preparing the exam after the course is released. It includes submitting exam questions and reporting the exam grades. The course coordinator needs to Upload Answer sheets after reporting the exam grades. The grades can vary between A to F. There is a special grade known as Fx. This grade means that the student has not passed the exam, but the submission was very close to the passing grade. In such a case, the course coordinator can Define and publish complementary Assignments for those students. The course coordinator needs to Correct Submissions and Report Grades in such a case. Note that this change will only be given to students once. 

After the course and its examination is over, the course coordinator shall evaluate the course. The evaluation starts by defining and publishing the evaluation form. After the evaluation is done, the course coordinator shall write and Submit the Course Evaluation.

\section{Result}\label{Sec:Result}

Among 24 master level students who registered for the exam, 20 students participated in \textit{Study 1} among which 13 students also participated in \textit{Study 2}.
In \textit{Study 1}, 9 and 11 students were male and female, respectively; 
while in \textit{Study 2}, 5 and 8 students were male and female, respectively.
The average age for students in both studies was around 34, who participated in the "Master's Programme in Open eGovernment" - which is a distance program. 
It is usual to have students with an industrial background in this course.

Students were familiar with BPMN process modeling languages, but they did not declare any prior experience on a declarative process modeling language.
Each student submitted one DCR and one CMMN model, so we collected different versions of their designs.
Here, we elaborate on how the process can be designed using these languages through two sample models.
We will elaborate on the study result afterward.

\subsection{Sample DCR model}

\begin{figure*}[t!]
	\vspace{-\baselineskip}
	\begin{center}
		\includegraphics[width=1\textwidth]{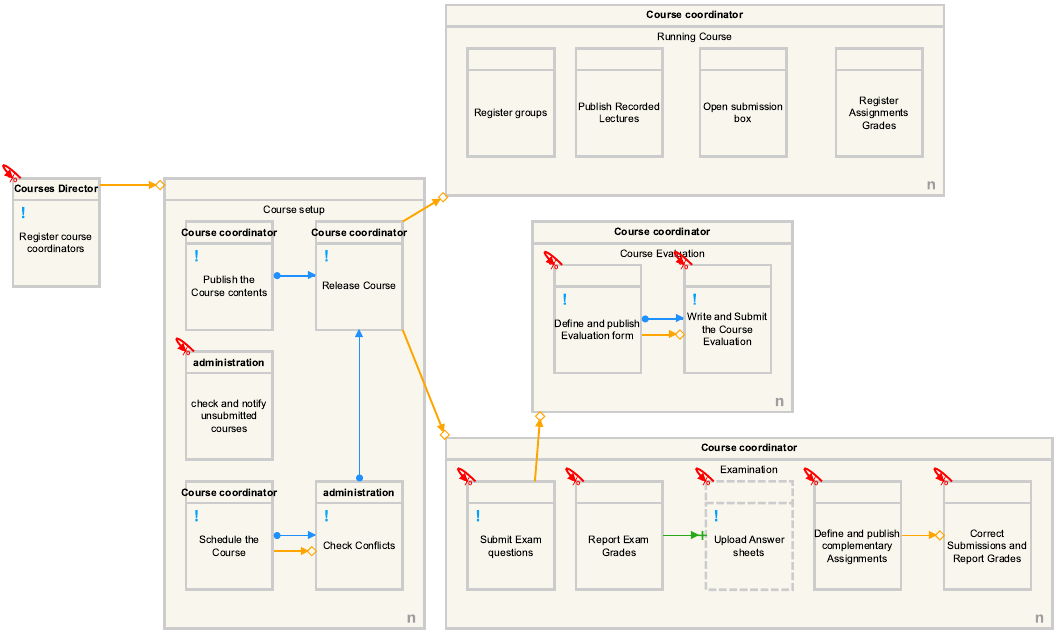}
		\caption{A sample DCR model for the given case description}\label{Fig:Exam_SampleModel_DCR}
		\vspace{-2\baselineskip}
	\end{center}
\end{figure*}

The case is modeled differently by each student, as there is the possibility to decide on the level of flexibility that a KiP model shall support.
\figurename~\ref{Fig:Exam_SampleModel_DCR} shows a sample DCR model for the given case description.

In this model, the start of the process is modeled to be rigid - meaning that the process has only one start point, i.e., \textit{Register course coordinators}.
The course director can perform this activity, and it is annotated as mandatory to apply more constraints for knowledge-workers to finish the process.
This activity excludes itself after execution, meaning that it can only be performed once! 
This constraint is not specified in the case description, and it can be considered as an extra control that the designer applied.
This constraint may cause a problem as the course director may want to assign another course coordinator later, which is not given in the case description.
As there might be many similar situations in the real world, it might be better if the process designer avoids applying extra control on models when designing KiPs.

The execution of \textit{register course coordinators} activity enables \textit{course setup}, which includes five activities.
The process model captures the case description. Here, the self-\textit{Exclude} relation applies extra control, which can be avoided.
After \textit{Release course} activity, \textit{Running course} and \textit{Examination} are enabled.
The \textit{Running course} is quite flexible to give freedom to the course coordinator to choose how to do the activities.
The \textit{Examination} includes five activities - all decorated with a self-\textit{Exclude} relation to limit the number of their execution.
All exclusions are fine as they are aligned with the case description, except \textit{Submit Exam questions}.
Again, the process is designed too rigid as it is expected for this node.

After \textit{Submit Exam questions} is performed, the \textit{Course Evaluation} will be enabled.
The self-\textit{Exclude} relations are not needed again here, as they make the process rigid.

\subsection{Sample CMMN model}

\begin{figure*}[t!]
	\vspace{-\baselineskip}
	\begin{center}
		\includegraphics[width=1\textwidth]{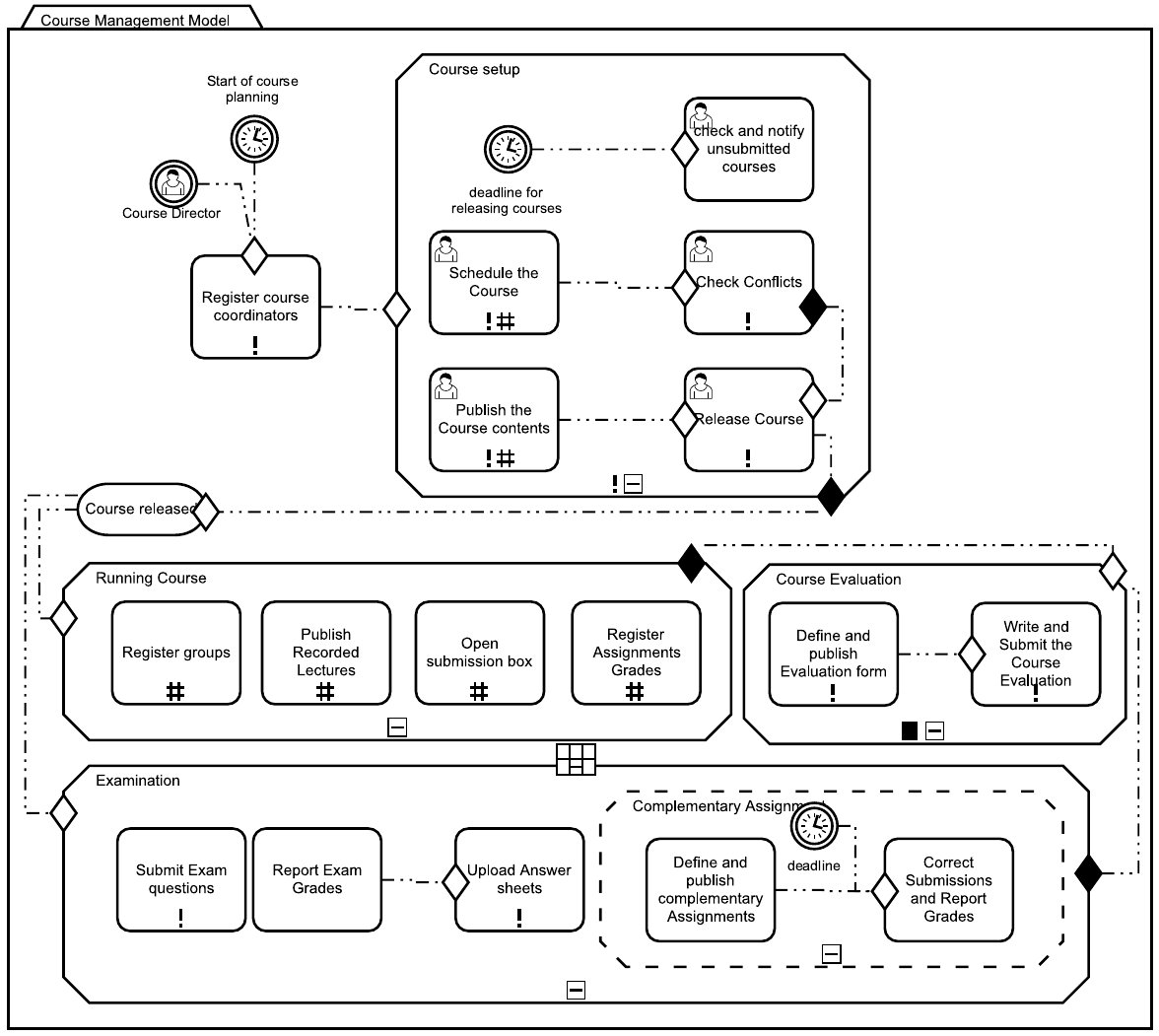}
		\caption{A sample CMMN model for the given case description}\label{Fig:Exam_SampleModel_CMMN}
		\vspace{-2\baselineskip}
	\end{center}
\end{figure*}

\figurename~\ref{Fig:Exam_SampleModel_CMMN} represents a sample CMMN model for the given case description.
In this model, every top-level task, stage, and milestone has a sentry as an entry criterion, which means that they are not enabled when the model is created - as their condition is not fulfilled.
Thus, the process starts when the \textit{Start of course planning} timer event occurs AND the course director starts \textit{Register course coordinators} task.
Note that if two events are related to an entry criterion sentry, both of them shall be fulfilled, which is equivalent to AND join in workflow-based models.
To show the OR relation, events shall be related to different entry criterion sentries, e.g., look at \textit{Release Course} activity.
To specify the resource that performs a task, one can relate the resource directly to the entry criterion.
However, this can make the model very complex.
Thus, \textit{DCR representation for resources seems better in terms of model simplicity}.

There is also one difference in how DCR and CMMN enable flexibility in terms of repeatable tasks/activities.
In DCR, an activity can be repeated unless explicitly limited. 
In contrast, \textit{a task is not repeatable in CMMN by default}. It needs to be decorated with \textit{Repition Rule} ($\#$). So knowledge-worker can repat the task.

When the course director performs the \textit{Register course coordinators}, the \textit{Course setup} stage would be enabled.
After the stage is completed, the \textit{Course released} milestone would be achieved.
This milestone enables \textit{Running Course} and \textit{Examination} stages.
Finally, the \textit{Course Evaluation} can be done to finish the process.

It is possible to analyze all students' submissions to explore how the case is modeled differently in these languages.
We skip this analysis as it is outside the scope of this paper.

\subsection{Perception Analysis}

\begin{figure*}[t!]
	\vspace{-\baselineskip}
	\begin{center}
		\includegraphics[width=1\textwidth]{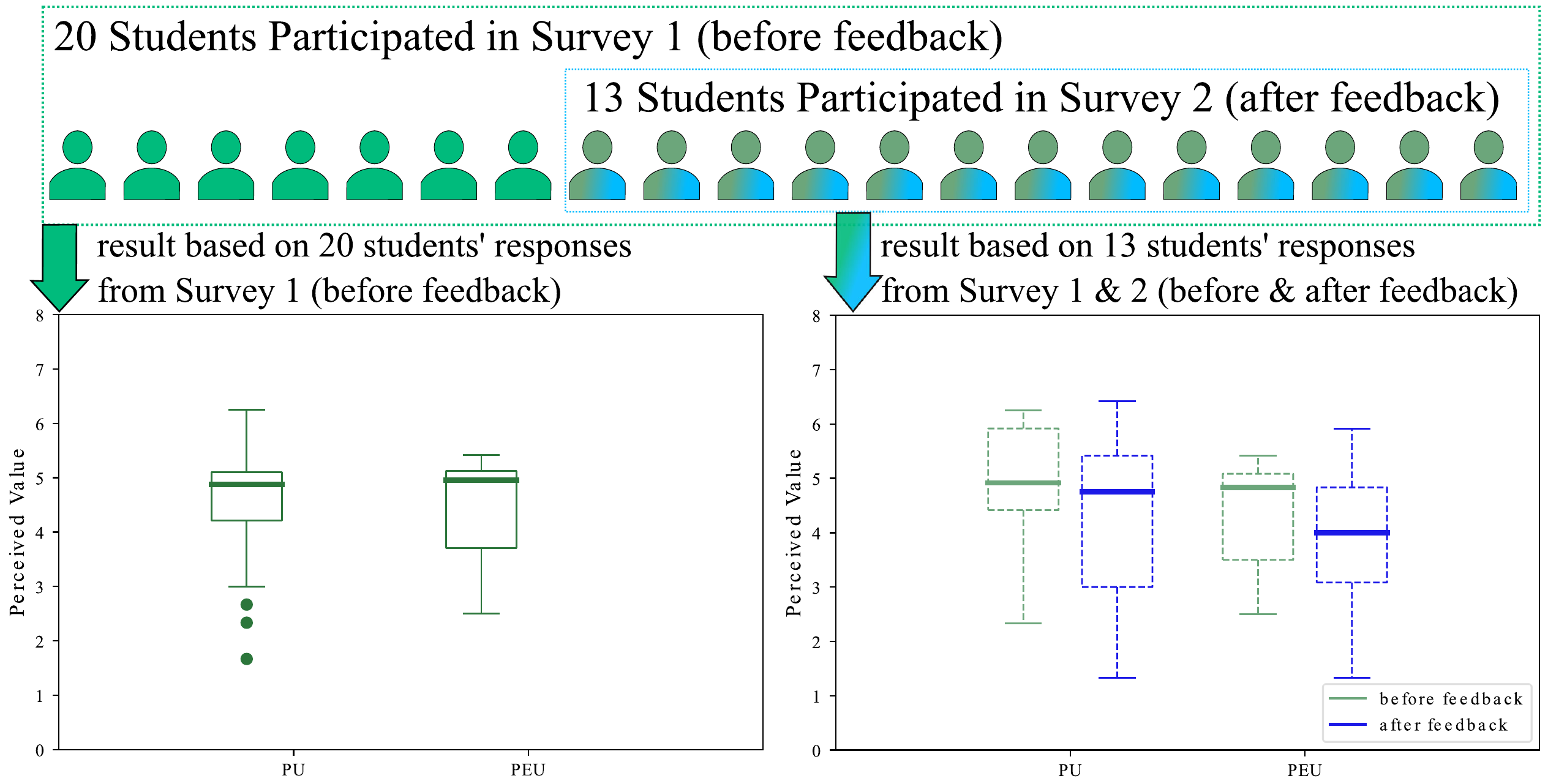}
		\vspace{-2\baselineskip}
		\caption{Aggregated Perceived Usefulness (PU) and Ease of Use (PEU)}\label{Fig:PUPEUBoth}
		\vspace{-2\baselineskip}
	\end{center}
\end{figure*}

\figurename~\ref{Fig:PUPEUBoth} shows how students perceived the utilized KiPs (i.e., DCR and CMMN) to be useful and easy to use through Box plots.
The figure is not specific for each of these languages.
The left- and right- side of \figurename~\ref{Fig:PUPEUBoth} shows the result of \textit{Study 1} and \textit{Study 2} representing the perception of students before and after recieving the feedback, respectively.
It worth reminding that some students from \textit{Study 1} did not participated in \textit{Study 2}, so the perception of students for \textit{Study 1} who participated in \textit{Study 2} is also demonstrated in the \textit{Study 2} through dashed box plot.

In \textit{Study 1}, represented by the left- sub-figure in \figurename~\ref{Fig:PUPEUBoth}), the median for both PU and PEU is around 5 (out of 7).
In \textit{Study 2}, represented by the right- sub-figure in \figurename~\ref{Fig:PUPEUBoth}), the median for PU before and after the feedback is around 5.
The difference is visible in the first and second quartiles, where some students lowered their scores for PU after receiving feedback.
The difference for PEU is even more, where the median is lowered by one after receiving the feedback.
We will check if the difference is significant later.



\begin{figure*}[b!]
	\vspace{-\baselineskip}
	\begin{center}
		\includegraphics[width=1\textwidth]{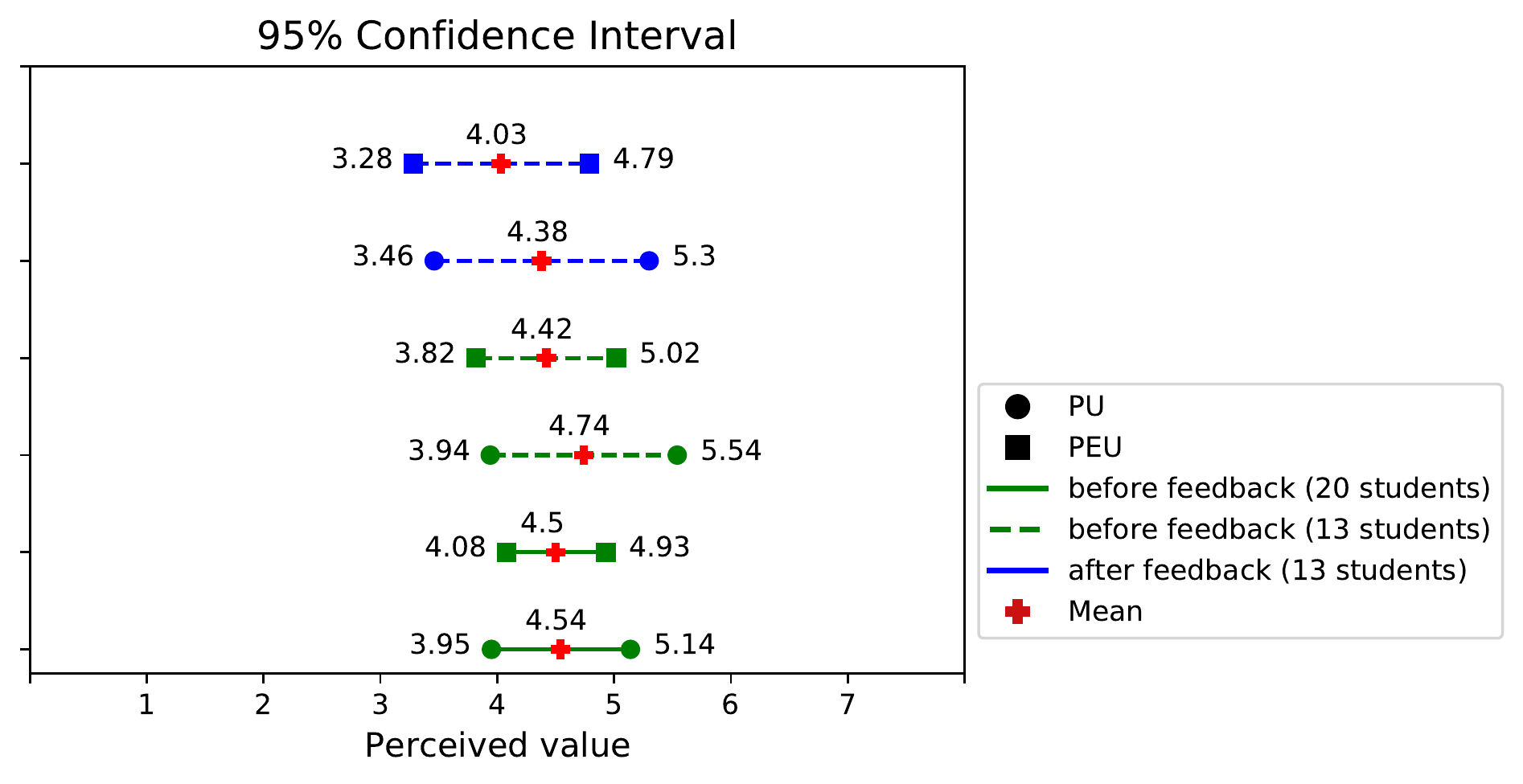}
		\caption{95\% Confidence Interval for the means}\label{Fig:CI_Aggregated}
		\vspace{-\baselineskip}\vspace{-\baselineskip}
	\end{center}
\end{figure*}

\begin{figure}[t!]
	\centering
	\subfigure[p-values for PU]{\includegraphics[scale=0.45]{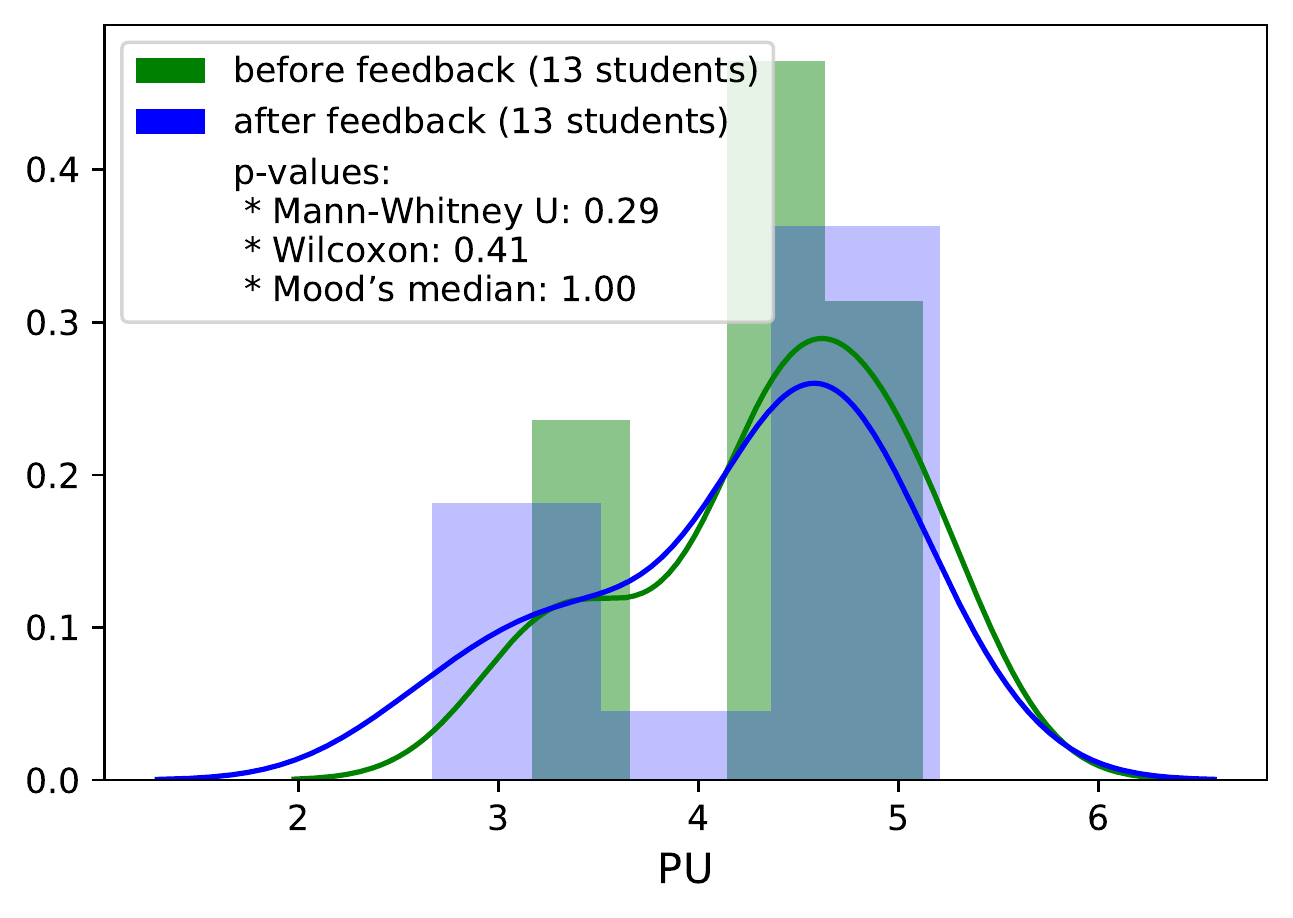}}\quad
	\subfigure[p-values for PEU]{\includegraphics[scale=0.45]{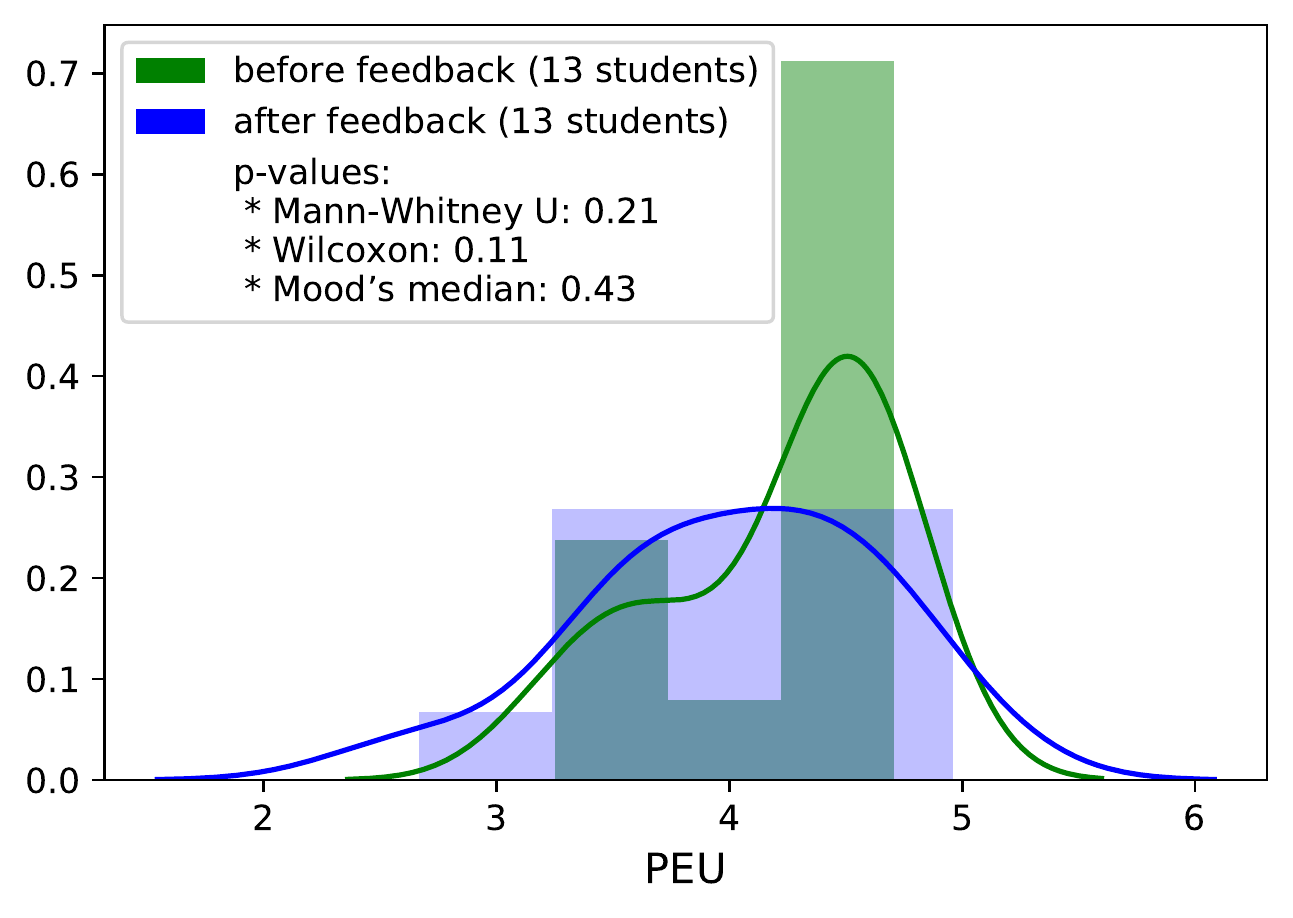}}
	\vspace{-1\baselineskip}	
	\caption{P-Values of PU and PEU for before and after feedback}
	\label{Fig:pvalues}
\end{figure}

\figurename~\ref{Fig:CI_Aggregated} shows the means and 95\% confidence interval for all measures that are demonstrated in \figurename~\ref{Fig:PUPEUBoth}.
The distribution of our data is not normal, so we cannot perform statistical tests like t-test to identify if the feedback had a significant difference in responses.
Therefore, we used three nonparametric statistical significance tests, i.e., 
Mann-Whitney U,
Wilcoxon Signed-Rank, and
Mood’s median tests.
We measured p-values based on these tests for PU and PEU in Study 2 for the responses that we received before and after the feedback.
\figurename~\ref{Fig:pvalues} shows the distribution of responses in these two cases in addition to the p-values.
The null hypothesis ($H_0$) is that the distribution of responses before and after the feedback are the same for both PU and PEU. 
The p-values are greater than 0.05, so we cannot reject the null hypothesis.
Thus, the feedback did not change the perceptions significantly.

\tablename~\ref{tab:PUPEULanguagues} shows the Cronbach's alpha result that we calculated per variable per language, where all values are above 0.7, which is generally considered as the accepted threshold.
The Cronbach's alpha for PU for both languages and PEU of DCR is quite high, i.e., above 0.9.
However, Cronbach's alpha for PEU of CMMN is not as high as others.

\tablename~\ref{tab:PUPEULanguagues2} shows the Cronbach's alpha result that we calculated per variable per language before and after feedback, where all values are above 0.7.
The Cronbach's alpha for all measures is quite high, i.e., above 0.9, except the Cronbach's alpha for PEU of CMMN before the feedback, which is 0.78.
It is worth mentioning that Cronbach's alpha for PU and PEU of both languages before the feedback is similar to their Cronbach's alpha for the whole population reported in \tablename~\ref{tab:PUPEULanguagues}.

\begin{table}[b!]
	\centering
	\vspace*{-0.2\baselineskip}
	\begin{tabular}{|l|c|c|}
		\hline
		& Perceived Usefulness (PU) & Perceived Ease of Use (PEU) \\ \hline
		DCR  & {\color[HTML]{005500} 0.97}                      & {\color[HTML]{005500} 0.94}                        \\ \hline
		CMMN & {\color[HTML]{005500} 0.97}                      & {\color[HTML]{005500} \textbf{0.70}}                        \\ \hline
	\end{tabular}
	\vspace{0.5\baselineskip}
	\caption{Cronbach Alpha for Study 1}
	\label{tab:PUPEULanguagues}
\end{table}

\begin{table}[b!]
	\centering
	\vspace*{-\baselineskip}
	\begin{tabular}{l|c|c|c|c|}
		\cline{2-5}
		& \multicolumn{2}{c|}{Perceived Usefulness (PU)}            & \multicolumn{2}{c|}{Perceived Ease of Use (PEU)}                            \\ \cline{2-5} 
		\multirow{-2}{*}{}         & Before Feedback             & After Feedback              & Before Feedback                      & After Feedback                       \\ \hline
		\multicolumn{1}{|l|}{DCR}  & {\color[HTML]{005500} 0.98} & {\color[HTML]{000055} 0.98} & {\color[HTML]{005500} 0.96}          & {\color[HTML]{000055} 0.97}          \\ \hline
		\multicolumn{1}{|l|}{CMMN} & {\color[HTML]{005500} 0.96} & {\color[HTML]{000055} 0.96} & {\color[HTML]{005500} \textbf{0.78}} & {\color[HTML]{00009A} \textbf{0.93}} \\ \hline
	\end{tabular}
	\vspace{0.5\baselineskip}
	\caption{Cronbach Alpha for Study 2}
	\vspace*{-.5\baselineskip}
	\label{tab:PUPEULanguagues2}
	\vspace*{-2\baselineskip}
\end{table}

\subsection{Discussion on biases and threats to validity}


First, we have used students as our test subjects instead of real process designers as explained and motivated in this paper.
Students are considered valid subjects in this area as these languages are new and are mostly unknown for practitioners outside.
Thus, students can be used to evaluate how these languages can be perceived by process designers, which is also used in related work such as ~\cite{pichler2011imperative,andaloussi2020understanding,sanchez2020supporting,andaloussi2019exploring,andaloussi2019exploringhybrid,andaloussi2018evaluating,haisjackl2016understanding,weber2009declarative}.
The use of students as subjects can weaken the causal relation for predicting if the artifact will be used in the future.
The fact that students belong to the same class and trained under the same process can also be considered as a learning bias.

Second, it shall be mentioned that students were familiar with BPMN business process modeling language, which may potentially impact their PU and PEU of declarative languages. From the author's perspective, this impact is unknown, and it will be interesting to evaluate if prior knowledge on workflow-based modeling language can have a positive or negative impact!

Third, feedback can impact the subjects' opinions as they can be used as positive or negative treatment.
However, the lack of feedback can result in under- or over- confidence biases.
We tried our best to use neutral wording~\cite{eberlein2011effects} to minimize this effect in this study.


\section{Conclusion}\label{Sec:Conclusion}

In this paper, we reported our study result on how CMMN and DCR has been perceived in terms of usefulness and ease of use.
The study is performed by applying the technology acceptance model, where we educated master level students with these languages over eight weeks by giving feedbacks to reduce perception biases. 
The students' perceptions are collected through two questionnaires as two sub-studies (studies 1 and 2).
We collected students' opinions before and after sending feedback on their final practice in the exam in study 1 and study 2, respectively. 
In total, twenty students participated in Study 1 among which thirteen students also participated in Study 2.

The study result indicates that both languages have an acceptable level for both perceived usefulness and ease of use.
The students' perceptions have changed a little before and after receiving the feedback.
We performed three nonparametric statistical significance tests, and the result indicates that the difference is not significant.
We also evaluated the reliability of responses using Cronbach's alpha.
The result showed an acceptable level of reliability in students' responses.

As future work, it is interesting to investigate how prior knowledge on workflow-based business process modeling can influence users' perception when learning declarative modeling languages.
It is also interesting to perform this study with more participants and different backgrounds.

\subsection*{Acknowledgement}
We appreciate all support that is provided by Morten Marquard from dcrgraphs.net, without which it was difficult to train the students and perform this study.

\bibliographystyle{abbrv}
\bibliography{References}

\end{document}